\documentclass[letterpaper, 10 pt, conference]{ieeeconf}  

\IEEEoverridecommandlockouts                              

\usepackage{graphicx}
\usepackage{booktabs}
\usepackage{multirow}
\usepackage{balance}

\title{\LARGE \bf
Sensitive Image Classification by Vision Transformers
}

\author{Hanxian He$^{1}$, Campbell Wilson$^{1}$, Thanh Thi Nguyen$^{1}$ and Janis Dalins$^{2}$
\thanks{*This work was supported by an Impact Grant through the Safer Children, Safer Communities program of Westpac Banking Corporation.}
\thanks{$^{1}$Hanxian He, Campbell Wilson and Thanh Thi Nguyen are with the AiLECS Lab, Faculty of Information Technology, Monash University, Melbourne, VIC 3800, Australia
        {\tt\small \{hanxian.he, campbell.wilson, thanh.nguyen9\}@monash.edu}}%
\thanks{$^{2}$Janis Dalins is with the AiLECS Lab, Australian Federal Police,
        Melbourne, VIC 3800, Australia
        {\tt\small janis.dalins@afp.gov.au}}%
}

\begin{document}

\maketitle
\thispagestyle{empty}
\pagestyle{empty}

\begin{abstract}

When it comes to classifying child sexual abuse images, managing similar inter-class correlations and diverse intra-class correlations poses a significant challenge. Vision transformer models, unlike conventional deep convolutional network models, leverage a self-attention mechanism to capture global interactions among contextual local elements. This allows them to navigate through image patches effectively, avoiding incorrect correlations and reducing ambiguity in attention maps, thus proving their efficacy in computer vision tasks. Rather than directly analyzing child sexual abuse data, we constructed two datasets: one comprising clean and pornographic images and another with three classes, which additionally include images indicative of pornography, sourced from Reddit and Google Open Images data. In our experiments, we also employ an adult content image benchmark dataset. These datasets served as a basis for assessing the performance of vision transformer models in pornographic image classification. In our study, we conducted a comparative analysis between various popular vision transformer models and traditional pre-trained ResNet models. Furthermore, we compared them with established methods for sensitive image detection such as attention and metric learning based CNN and Bumble. The findings demonstrated that vision transformer networks surpassed the benchmark pre-trained models, showcasing their superior classification and detection capabilities in this task.

\end{abstract}

\section{INTRODUCTION}

The rapid development of digital technology has significantly changed people's lives, allowing more young people, including children, to engage in online activities easily. With the widespread popularity of platforms like TikTok, Instagram, and similar video and live streaming applications, coupled with advancements in deepfakes and AI technologies, there has been a growing proliferation of explicit content, especially during the pandemic. The need for effective and reasonable classification and management of explicit content has become increasingly urgent. Unfortunately, only 37\% of these technology companies have implemented measures to detect grooming online, according to a survey by the WeProtect Global Alliance \cite{weprotect_global_alliance_findings_2021}. Telegram received severe criticism for not taking sufficient measures to address the large volume (around 100,000 instances) of pornographic content involving women generated using the DeepNude AI tool on its platform in 2020 \cite{Badiei2021}. 

The use of AI technology to intervene in pornography detection is a current trend \cite{lee_detecting_2020}. First, addressing online grooming and live streaming of sexual material requires more direct online pornography detection techniques. Secondly, the generation of pornography material through deepfake technology necessitates advanced AI detection methods. The third point is that research on explicit content, especially child sexual abuse material (CSAM), requires more refined and accurate methods for rapid analysis. Furthermore, it is crucial to identify sensitive content from vast amount of information and detect and prevent its dissemination to minimize secondary harm. This requires specialized pornographic automatic detection and classification technologies that can provide quantifiable metrics and classification standards for criminal investigation and law enforcement while reducing direct exposure to sexual content. 

Currently, the classification of pornographic image material is primarily focused on two directions: one is simply distinguishing whether it is pornographic material, and the other involves classifying and grading different levels of pornographic material based on its content. However, in practical operations, establishing a clear and unified classification and grading standard, whether for image and video content rating, content moderation, or criminal-related pornography classification, proves to be challenging.

Due to the sensitivity of pornographic image material and the influence of cultural, religious, and political factors in different countries, the management of pornographic material varies significantly. Even for research purposes, the creation and sharing of relevant datasets face substantial limitations, resulting in a scarcity of benchmark data in this domain. Moreover, the classification research related to pornographic images is more prevalent in various countries' police systems, employing traditional machine learning or CNN-based methods. Only recently has there been growing attention to methods based on vision transformers, which were originally developed for natural language processing tasks, and now also for image recognition tasks.

In this study, we conducted a performance comparison between different vision transformer models and traditional methods in pornographic image detection tasks. The structure of this paper is organized as follows. First, a literature review is presented, followed by the methods applied in the paper. Then, comparison experiments on pornography and porn-indicative image detection are provided. Lastly, the results are analyzed and the limitations of this study and potential research directions are mentioned in the conclusion section. 

\section{RELATED WORK}

\subsection{Pornography Datasets}

Table \ref{tab:dataset} lists the majority of pornographic datasets widely used in research. Among them, adult content image (ACI) \cite{connie2017smart}, P2datasetFull \cite{figshare_20k_nodate}, and Pornography-2M \cite{gangwar_attm-cnn_2021} are datasets in which images are labeled only based on whether they are explicit or not. Meanwhile, Nudenet goes a step further by providing detailed annotations based on the exposure of different body parts on top of the two-class labeling. Similarly, data sets such as LSPD, sexACT, and AIIA-PID4 have further detailed annotations that cover additional information. The LSPD dataset provides 93,810 labeled instances among 50,212 images. Among these, pornographic images are primarily sourced from adult websites, while non-pornographic images are obtained by searching 250 categories of keywords on Google. For each keyword, approximately 1,000 images were downloaded. The pornographic images are further annotated using polygon masks for four private sexual organs: breasts, male and female genitals, and anus. The sexACT dataset was created by
 \cite{oronowicz-jaskowiak_description_2022} from images downloaded from Google by searching through keywords. Although 5,000 images of children sexual abuse (CSA) provided by Spanish Police and 1,112,531 images of adult from the Pornography-2M dataset are mentioned by \cite{gangwar_attm-cnn_2021} from the Spanish Group for Vision and Intelligent Systems (GVIS), the dataset is not available for general usage.
 
\begin{table}\footnotesize
  \caption{Image-based Sensitive Datasets}
  \centering
  \begin{tabular}{c c c c}
    \toprule
    Name & Paper & Year & Labels \\
    \midrule
      LSPD & \cite{phan2022lspd} & 2022 & porn, normal, sexy, hentai, 
    \\
         & &  &  drawings, female/male genital
    \\
            & &  &  female breast, anus
    \\
    \hline
       Nudenet & \cite{notai-tech_nudenet_nodate} & 2019 & nude or not
    \\
    \hline
        ACI & \cite{connie2017smart} & 2017 & nude or not
    \\
    \hline
      AIIA-PID4 &\cite{detectingpornrois}& 2013 & bikini, porn, skin, non-skin
      \\
      \hline
      sexACT &\cite{oronowicz-jaskowiak_description_2022}& 2022 & AB/DL, sexual activity,
      \\
      & & & nude women, dressed women, 
       \\
      & & &sexual activity-spanking
      \\
      \hline
      P2datasetFull &\cite{figshare_20k_nodate}& 2020 & nude or not
     \\
     \hline
      Pornography-2M &\cite{gangwar_attm-cnn_2021}& 2021 & nude or not
     \\
    \bottomrule
  \end{tabular}
  \label{tab:dataset}
\end{table}

Due to cultural, religious, and other factors, definitions of explicit and non-explicit content vary. Additionally, definitions of pornography can evolve rapidly over time. As societal norms, cultural perspectives, and legal frameworks change, the understanding and categorization of explicit content may undergo significant transformations. Therefore, it is necessary to introduce models capable of detecting pornography content and details while incorporating a separate category for porn-indicative content that falls between safety and explicitness. In this paper, we define three classes (categories) involved in the detection of pornography images, which is shown in Table \ref{tab:tb1}. Building upon this foundation, we can further classify pornography materials into more specific categories based on the varying laws and cultural customs of different countries. These subcategories can serve different purposes, such as law enforcement efforts, and allow for a more nuanced classification and handling of sensitive materials.

\begin{table}\footnotesize
  \caption{Categories of Sexual Abuse Materials in This Paper}
  \centering
  \begin{tabular}{@{}lc@{}}
    \toprule
    Guidelines & General Category \\
    \midrule
    {Clean} & {Images with no sex-related} \\ & {or suspicious sex content}
    \\
    \hline
    {Porn Indicative} & {Images with indicative or suspicious sex content} \\ & {but with no obvious sexual organ exposure}
    \\
    \hline
    {Pornography/Sensitive} & {Pornographic/sexual images} \\ & {with human beings and sex content}
     \\
    \bottomrule
  \end{tabular}
  \label{tab:tb1}
\end{table}

\subsection{Pornography Detection Tools}
Pornography detection has been widely used in content moderation, sensitive content detection, scene genre classification, and so on.  AI algorithms allow for more sophisticated analysis and detection of pornography materials. The automatic pornography detection tool NuDetective \cite{de_castro_polastro_nudetective_2010} developed by the Brazilian Federal Police is one of the most prominent binary machine learning-based forensic pornography detection tools. Of all these pornography detection models based on ResNet, the Yahoo Open Not Safe For Work (NSFW) model \cite{noauthor_open_2023} developed in 2016 is the most representative. The Yahoo Open NSFW model is a binary ResNet-50 model finetuned from the ImageNet dataset for NSFW classification, but the probability scoring and threshold setting design hinder the wider application of this model. The Majura labeling schema proposed by \cite{dalins_laying_2018} also adopted pre-trained VGG and ResNet18 models for child exploitation material (CEM) labeling. Similar open-sourced sensitive image and video scene detection tools include the LAION safety toolkit\footnote{https://github.com/LAION-AI/LAION-SAFETY}, the PysceneDetect\footnote{https://github.com/Breakthrough/PySceneDetect}, the ffsubsync\footnote{https://github.com/smacke/ffsubsync}, and the MoviePy\footnote{https://github.com/Zulko/moviepy}. Among them, the LAION safety toolkit is an NSFW detector trained on the LAION image dataset. The other three, however, are detection software based on videos or video clips.

\subsection{Pornography Detection Models}

Different from pornography detection and rule-based traditional machine learning methods, utilizing deep learning to automatically extract key features for sensitive (pornography) content recognition allows for more comprehensive learning of factors that influence sensitive material classification, including pornography detection through feature maps. For AI model-based pornography detection models, ResNet and attention model methods are the two most popular types of deep learning networks deployed in NSFW or CSAM material detection. The work by \cite{oronowicz-jaskowiak_description_2022} utilized the ResNet models from Fast.ai and PyTorch libraries for NSFW classification. In that paper, the ResNet models including the ResNet152 and ResNet101 demonstrated better performance than the VGG19, VGG16, SqueezeNet 1.1, and SqueezeNet~1.0 models in the classification of 9 types of pornographic materials. It also showed that the exclusion of any individual category did not significantly influence the overall classification accuracy with the ResNet152 model, which means the classification system is not a good choice. The classification relies on conditional judgment-based labeling, which falls short of achieving fully automated sensitive content detection.

Attention-module-based methods or transformer models have become increasingly popular in comparison with pre-trained models like ResNet18, primarily due to their higher accuracy, albeit at the cost of increased computation. The concept of attention models (AM), initially introduced for machine translation by \cite{bahdanau_neural_2014} in 2014, has now become a prominent concept in the field of neural networks for a wide range of applications, including computer vision \cite{wang_survey_2016}. Attention models incorporate this concept of relevance by dynamically allowing the model to focus its attention on specific parts of the input that are most valuable for performing the task at hand effectively. In sensitive image detection, transformers were first introduced for action recognition in video clips. A significant amount of research has been proposed based on the Kinetics \cite{kay2017kinetics} and SSv2 \cite{goyal2017something} datasets, both of which consist of human action video clips. Simultaneously, there is a growing popularity in utilizing transformers for multimodal video analysis \cite{multimodaltransformer}. This includes models such as CLIP-like \cite{lin2022frozen}, ViT or ViT-like \cite{tong2022videomae,bertasius2021space,likhosherstov2021polyvit,girdhar2023omnimae,girdhar2022omnivore,xiong2022m,bachmann2022multimae}, and models similar to the Swin transformer \cite{liu2022video,wang2022bevt}. An attention model with a metric learning module named AttM-CNN was introduced in \cite{gangwar_attm-cnn_2021}  to realize child sexual abuse (CSA) content detection in 2021. In their work, the AttM-CNN module has been applied to both pornography image detection and age detection networks step by step. According to the score achieved by two steps, the image is classified into different CSA content categories by a binary classification system. The combination of the Inception module as well as the attention module makes this model outperform other CSA detection systems, including the Nudetective and Yahoo Open NSFW systems. However, the classification based on scoring rules makes it challenging to deploy this model to other cases. Additionally, the related models and data are not public.

\section{BACKGROUND}
\label{sec:Background}

Most classical neural sequence transduction models including the transformer models follow the same encoder-decoder structure \cite{vaswani2017attention}. The transformer model primarily consists of the encoder and the decoder two parts. In the encoder, multiple stacked layers are utilized, and each layer comprises a self-attention sub-layer and a feed-forward neural network sub-layer. On the other hand, the decoder has a similar structure with stacked layers, but each layer includes three sub-layers. These sub-layers consist of a self-attention sub-layer, a feed-forward neural network sub-layer, and an additional encoder output attention sub-layer for reinforcement. Furthermore, both the encoder and decoder have a process of residual connections and normalization after each sub-layer.

\subsection{Vision Transformer}

\begin{figure}[thpb]
\centering
\includegraphics[width=1\linewidth]{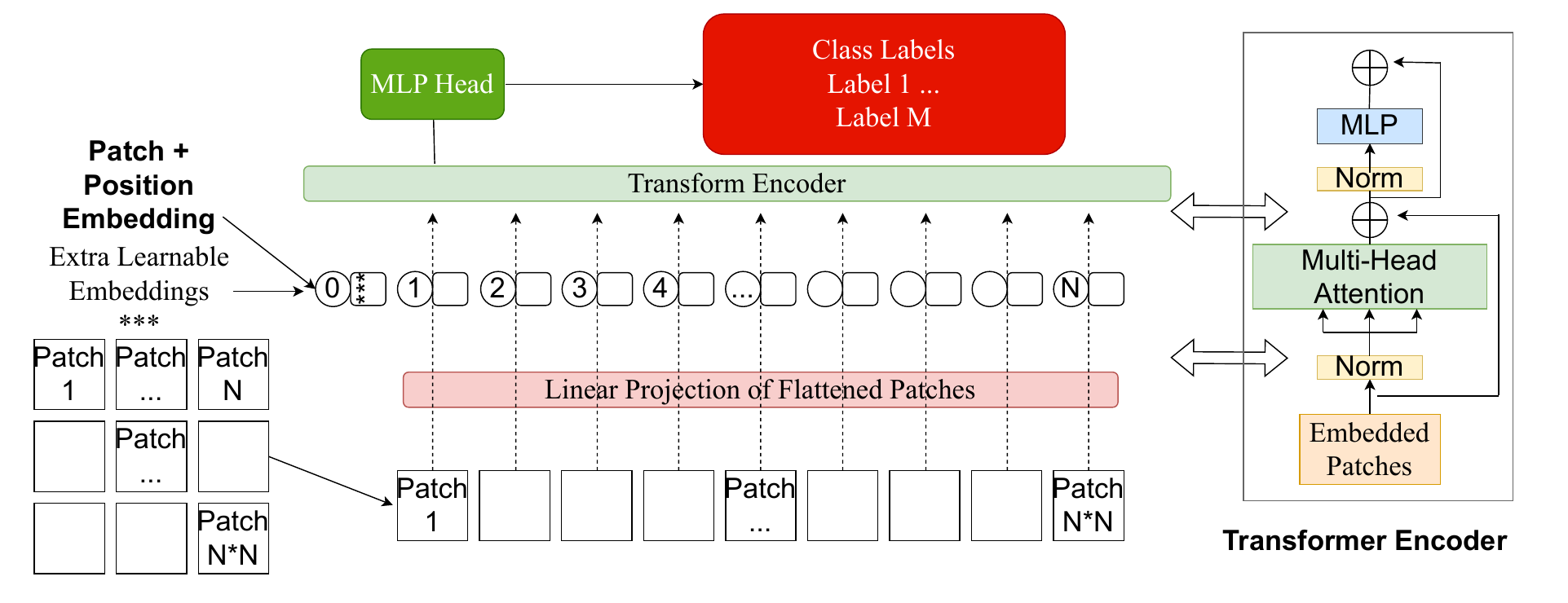}
\caption{A structure of the ViT transformer module (redrawn from \cite{dosovitskiy2020image}).}
\label{fignew2}
\end{figure}

The vision transformer, commonly known as ViT, is a model designed for image classification tasks. It utilizes a transformer-based architecture that operates on image patches. The input image is divided into patches of a fixed size. Each patch is linearly embedded, incorporating its visual features, and position embeddings are introduced to capture spatial information. The resulting sequence of embedded vectors is then passed through a standard transformer encoder. To enable classification, a conventional technique is employed, which involves appending an additional learnable ``classification token" to the sequence. This token serves as a representative element that summarizes the information from the image patches and aids in making the final classification decision. Using the attention mechanism of the transformer and the positional embeddings, the ViT achieves effective image classification capabilities. The structure of the ViT transformer module is provided in Fig. \ref{fignew2}.

\subsection{Swin Transformer Model}
\label{subsec:SWin}
\begin{figure}[thpb]
\centering
\includegraphics[width=0.48\textwidth]{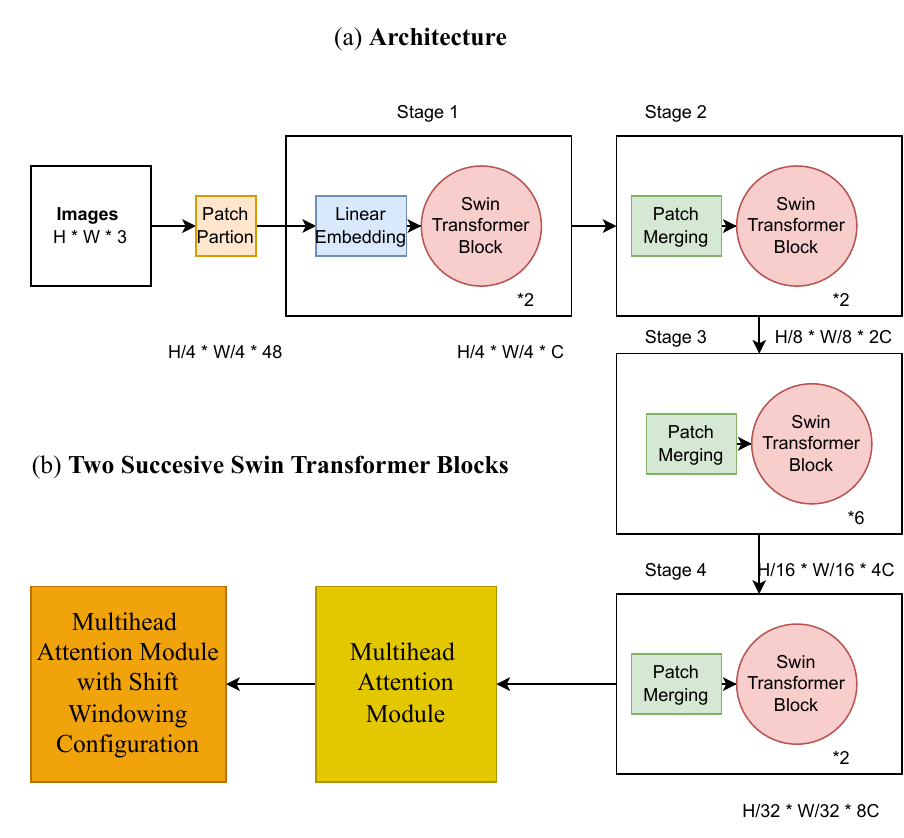}
\caption{A structure of the Swin transformer model (redrawn from \cite{liu2021swin}). In a patch partitioning scheme with a 4 $\times$ 4 size, the approach generates feature vectors at the size of 48 from the input image. These vectors are obtained through a linear embedding layer and then processed by a Swin transformer block, resulting in a feature layer with dimension C.}
\label{fig4}
\end{figure}

The Swin transformer is a hierarchical vision transformer using ``shifted windows", introduced in \cite{liu2021swin}. Unlike traditional transformers that fixedly process input sequences, the Swin transformer operates on non-overlapping patches of an image. It adopts a hierarchical design by stacking multiple stages, where each stage contains a set of transformer blocks. In each stage, a shifted window mechanism is employed to capture both local and global contextual information efficiently. This mechanism allows for the capture of long-range dependencies while maintaining computational efficiency.  The structure of a Swin transformer is provided in Fig. \ref{fig4}. The linear embedding and swin transformer block process are repeated in the second, third, and fourth stages. In each stage, the features from the patch partitions are merged before being input to another Swin transformer block, producing a new output with a different dimension.

\subsection{HiLo Attention Model}
\label{subsec:LITv2}

\begin{figure}[!htbp]
\centering
\includegraphics[width=1\linewidth]{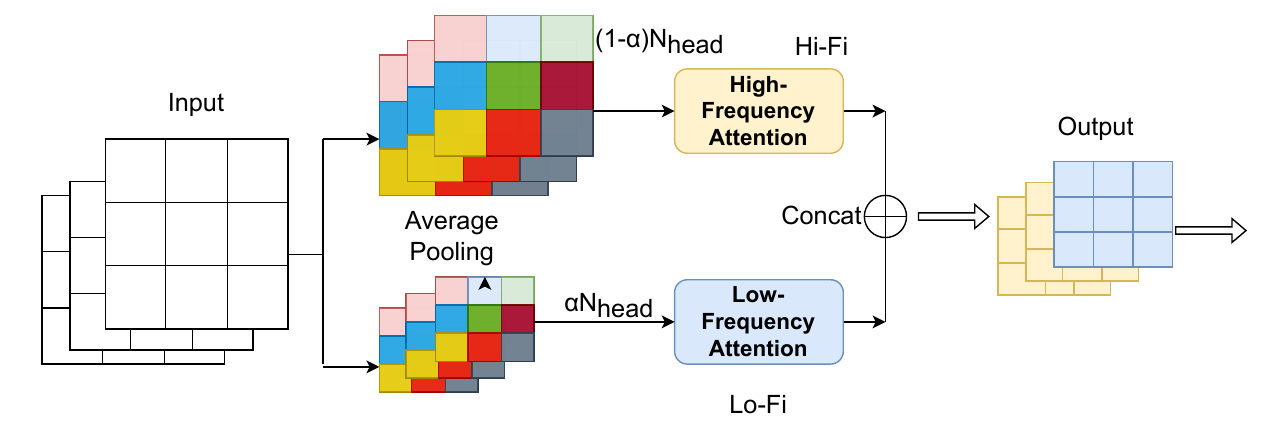}
\caption{The Structure of the Hi-Lo attention model LITv2 (redrawn from \cite{pan_fast_2023}). $H_h$ represents the total number of attention heads. $\alpha$  represents the ratio of low-frequency attention heads to total attention heads.}
\label{fig5}
\end{figure}

LITv2, also named improved fast vision transformers with HiLo Attention, is a model proposed by \cite{pan_fast_2023}. The HiLo attention mechanism draws inspiration from the observation that different frequencies can capture image features from diverse granularity, i.e., high frequencies for local fine details and low frequencies for global features. This facilitates capturing global contextual information and structural dependencies across the entire image. By separating the attention heads based on frequency characteristics, the HiLo attention mechanism enables the model to capture both local fine details and global structures effectively. This disentanglement approach improves the ability of the attention mechanism to represent different frequency patterns, leading to enhanced performance in speed and memory consumption. The structure of the LITv2 model is provided in Fig. \ref{fig5}.

\section{EXPERIMENTS}
\label{sec:experiments}

\subsection{Datasets}
\label{subsec:data}
In most CSAM classification studies, images or videos are often simply categorized as either pornography or non-pornography. However, in real-world scenarios, the definition of pornographic images can vary significantly due to factors such as country, culture, religion, and other aspects. To provide a more accurate definition of pornographic and nonpornographic images, in our article, we defined certain images that are suspected to contain pornographic content as suspicious pornography, i.e., porn-indicative. The definition of each category is provided in Table \ref{tab:tb1}. Our experiments in this study employed three datasets, namely 2-class P2, 3-class P2, and ACI, as described in detail below. 

The process of creating the P2 dataset consisted of three stages. Basically, the idea is to supplement data based on the original P2datasetFull dataset introduced in \cite{figshare_20k_nodate}. The P2datasetFull is a public dataset with pornography and clean images collected online in three subfolders of training, testing, and validation. However, this dataset consists of a large number of invalid images. For the purpose of creating a benchmark dataset for pornography image classification, we gathered samples from various public resources, including Reddit, ImageNet, and the P2datasetFull dataset. The first stage involved extracting photos indexed on Reddit and ImageNet. Pornographic and indicative images are mostly collected from the Reddit NSFW dataset. Meanwhile, neutral/clean images of human beings are selected from the ImageNet dataset. The second stage involved the automatic deletion of photos that were invalid or could not be read. The third stage involved the manual selection of images to exclude blurred and duplicated images, as well as organize samples into subfolders based on their labels and content, namely pornography, porn indicative, and clean.  Following this process, we were able to create a benchmark pornography dataset, allowing us to evaluate and compare the performance of different classifiers in image classification tasks. The images in the P2 dataset are finally standardized at a resolution of 256 $\times$ 256 pixels. We name the three-class dataset created as \emph{3-class P2} and the reorganized two-class dataset without the indicative category as \emph{2-class P2}.

To compare with the state-of-the-art research work, another 2-class sensitive dataset named ACI \cite{connie2017smart} is also included in our experiments. The ACI dataset is inspected as clean and organized and will be utilized directly in this paper. The images in the ACI dataset are all standardized to a resolution of 128 $\times$ 128 pixels. 

We acknowledge the ethical concerns around the use of adult sexual data obtained through the reuse of scraped datasets and do not in any way condone this methodology more broadly. However, we have made use of this dataset in the interests of facilitating comparison of our results with other work in this area.

\subsection{Experiment Settings}
\label{subsec:exp_settings}
In previous related work, \cite{dalins_laying_2018} utilized the ResNet18 architecture, which was pre-trained with the Google Image dataset, to address the task of CSAM classification. Specifically, leveraging the features learned from a large and diverse dataset, the pre-trained ResNet18 model demonstrated its effectiveness in accurately identifying and classifying CSAM material. In this paper, we compared the pre-trained ResNet18 methods with several popular image transformer-based models, i.e., ViT, DeiT, Swin and LITv2 models with various alpha settings.

\begin{table}[]
    \caption{ The composition of the P2 dataset with 3 classes}
    \centering
    \begin{tabular}{c c c c}
       \hline
        P2 dataset & Training & Validation & Testing\\ [0.1ex]
       \hline\hline
       Clean & 3328 & 3328 & 3301 \\
       \hline
       Pornography & 2432 & 2304 & 2387\\
       \hline
       Porn-indicative & 512 & 640 & 450 \\
       \hline    
    \end{tabular}
    \label{tab:t2}
\end{table}

\begin{table}[]
    \caption{ The composition of the ACI dataset}
    \centering
    \begin{tabular}{c c c c}
       \hline
        ACI dataset & Training & Validation & Testing\\ [0.1ex]
       \hline\hline
       Clean & 27,984 & 6104 & 6131 \\
       \hline
       Pornography & 28930 & 6092 & 6065 \\
       \hline    
    \end{tabular}
    \label{tab:ACI}
\end{table}

\subsection{Results and Analysis}
\label{subsec:results}

For most vision transformer models, three sizes of models are provided on their official website according to the flops, the memory required, and the number of parameters, i.e., base, medium, and small models. In general, the term ``base model" typically implies a larger model with more parameters and computational complexity compared to a ``medium model," and a ``medium model" is larger than a ``small model." A base model tends to have greater learning capacity during training due to its increased parameters and layers. However, a small model is more efficient and memory-saving. The medium model makes a trade-off between learning capacity, computational efficiency, and memory requirements. Here, we take LITv2 as an example to compare the differences between various models sizes in terms of the dimension of the embeddings, the number of heads, the depths of attention, as well as the number of parameters in Table \ref{tab:experiment1}. Taking the LITv2-base as a reference, the dimension of embeddings is 128, the number of attention heads is [4, 8, 16, 32], and the depth of attention modules is [2, 2, 18, 2]. The input size of all these three models is 224 $\times$ 224 pixels.
In Table \ref{tab:experiment2}, we also provide the comparison of the most popular vision transformers, including ViT, DeiT, and Swinv2. Here, ViT-base is equivalent to the ``small" models in other vision transformers, as its naming convention follows ``base", ``large", and ``huge", which corresponds to ``small", ``medium" and ``base" in other models.

\begin{table}[b]\footnotesize
    \caption{Experiment settings in LITv2 models. Here DoE is short of Dimension of Embeddings. The local patch size of these models is [0, 0, 2, 1] }
    \centering
    \begin{tabular}{c c  c c }
    \hline
        Model   & DoE & Number \& Depth of Attentions & Params (M) \\
        \hline
        LITv2-base & 128 & [4, 8, 16, 32], [2, 2, 18, 2] & 87 \\
        LITv2-medium & 96 & [3, 6, 12, 24], [2, 2, 18, 2] & 49\\
        LITv2-small& 96 & [3, 6, 12,24], [2, 2, 6, 2] & 28\\
    \hline
    \end{tabular}
    \label{tab:experiment1}
\end{table}

Limited to the computation resources, we select the small models to compare their performance on our datasets. As mentioned in Section \ref{subsec:LITv2}, the composition of high-frequency and low-frequency attention is a vital factor in the representation of images. To compare the model performances with different ratios of high-frequency and low-frequency attention, we also test the LITv2 model with the ratio values of 0.1, 0.4, and 0.9, respectively, for small LITv2 model configurations in our case. The maximum accuracy reported in the validation data sets for the models trained from scratch (with 300 training epochs) is reported in Table~\ref{tab:result1}. The accuracy reported on the test data set for these trained models is reported in Table \ref{tab:result2}.

\begin{table}[b]\footnotesize
    \caption{Experiment settings for selected ViT models pre-trained on ImageNet-1k from the hugging-face website.}
    \centering
    \begin{tabular}{c  c  c}
    \hline
        Model   & Checkpoints Pre-trained on ImageNet-1k  &Params (M)\\
        \hline
        ViT-base & vit-base-patch16-224 & 86\\
        DeiT-small & deit-small-patch16-224 & 22\\
        Swinv2-small & swinv2-small-patch4-window8-256 & 30\\
    \hline
    \end{tabular}
    \label{tab:experiment2}
\end{table}

\begin{table}[b]\small
    \caption{The maximum accuracy on the validation dataset after 300 epochs training from scratch for each model.}
    \centering
    \begin{tabular}{c c   c c }
    \hline
        Val Accuracy    & P2 (2-class)  & P2 (3-class) & ACI (2-class)  \\
        \hline
        LITv2-S (0.1) & 0.9531 &  0.8406 & 0.9865\\
        LITv2-S (0.4) & 0.9315 &  0.8444 & 0.9867\\
        LITv2-S (0.9) & 0.9659 &  0.7920 & 0.9857\\
        Swin-S & 0.9045 & 0.7717 & 0.9696\\
        DeiT-S & 0.9027 & 0.7971 & 0.9665 \\
    \hline
    \end{tabular}
    \label{tab:result1}
\end{table}

\begin{table}[b]\small
    \caption{The test accuracy on the test dataset after 300 epochs training from scratch for each model.}
    \centering
    \begin{tabular}{c c   c c }
    \hline
        Test Accuracy    & P2 (2-class)  & P2 (3-class) & ACI (2-class)  \\
        \hline
        LITv2-S (0.1) & 0.7619 &  0.8395 & 0.9809\\
        LITv2-S (0.4) & 0.7597 &  0.8531 & 0.9636\\
        LITv2-S (0.9) & 0.7714 &  0.8011 & 0.9552\\
        Swin-S & 0.8833 & 0.7493 & 0.9672\\
        DeiT-S & 0.8639 & 0.7532 & 0.9108 \\
    \hline
    \end{tabular}
    \label{tab:result2}
\end{table}

From Table \ref{tab:result1}, it can be observed that, even during the training phase, the classification results for the P2 3-class datasets, including the porn indicator category, are significantly lower compared to the classification with only two classes: pornography and clean data. This is in part due to the limited amount of data in the porn-indicative category, resulting in a clear bias. Additionally, the introduction of porn-indicative data diminishes the interclass differences between porn and clean categories, affecting the classification results. Simultaneously comparing the training accuracy between the P2 dataset and ACI, all vision transformer models demonstrate higher training accuracy on the ACI dataset, which has more training data. Additionally, overall, the LITv2 model outperforms the Swin-S and Deit-S models. Meanwhile, the Swin-S and DeiT-S models exhibit comparable performance across all three datasets.

Examining Tables \ref{tab:result1} and \ref{tab:result2}  together reveals that the classification results in independent test data for models trained on the two-class dataset noticeably decrease compared to their performance on the validation data. However, on datasets that include the porn-indicative category, the models consistently demonstrate stable performance. Specifically, on the 2-class P2 dataset, LITv2 models with different ratios exhibit a significant reduction in classification accuracy, approximately 17\% to 20\%, while the Swin and DeiT models also show a decrease in accuracy, about 2\% to 4\%. For the ACI dataset, the small difference between the accuracy of the validation and the accuracy of the test can be attributed to several factors. Firstly, the ACI dataset has a larger training data size compared to the P2 dataset. Secondly, the distribution of data among the train, validation, and test sets in the ACI dataset is more consistent. This consistency in data distribution ensures that the model is trained on a representative sample of the data, allowing it to generalize well to unseen test data. The LITv2-S models with different ratios do not exhibit significant differences in the three datasets. Meanwhile, Swin and DeiT models outperform LITv2 models on the P2 2-class dataset. However, in the P2 3-class dataset, LITv2 models outperform both Swin and DeiT models.

To further analyze various vision transformer models based on different mechanisms, we performed experiments comparing three pre-trained models mentioned in Table \ref{tab:experiment2} with the re-trained ResNet18 model \cite{dalins_laying_2018} across different datasets and the results are provided in Table \ref{tab:result3} and \ref{tab:result4}. The comparison reveals that, despite the occurrence of overfitting, the accuracy of all models is significantly higher than that of a ResNet18 model retrained on ImageNet-1K. Moreover, due to the utilization of pre-trained models based on ImageNet-1K, the final test accuracy is considerably higher compared to models trained from scratch. Meanwhile, based on the data in Table \ref{tab:result3} and \ref{tab:result4}, models such as ViT, Deit and Swin, fine-tuned from saved checkpoints pre-trained on ImageNet-1K, achieved an accuracy greater than 95\% on the P2 2-class dataset, surpassing the performance of AttM-CNN \cite{gangwar_attm-cnn_2021}, which achieved an accuracy of 92.72\%. Additionally, despite the influence of input image resolution and the lack of a unified data comparison, the overall accuracy is comparable to the reported average accuracy in this article \cite{nuditycomparision}. Meanwhile, we also compared our results with the Bumble commercial content moderation model \cite{the_bumble_tech_bumbles_nodate}, which is adopted from the EfficientNetv2~\cite{tan2020efficientnet}. Due to the lack of threshold details in the Bumble model, we selected 20\% as the boundary threshold between clean and pornography categories after a series of threshold comparisons. This yielded overall accuracy of 89.27\% and 89.87\% on the P2 and ACI datasets, respectively. Compared to the Bumble model, our model's performance is superior. However, among the various vision transformer models, there is little difference in accuracy.

\begin{table}[b]\small
    \caption{The validation accuracy on the validation dataset after 50 epochs fine-tuning from checkpoints pre-trained on ImageNet-1K for each model.}
    \centering
    \begin{tabular}{c c   c c }
    \hline
        Val Accuracy    & P2 (2-class)  & P2 (3-class) & ACI (2-class)  \\
        \hline
        ViT-base & 0.9957 &  0.9949 & 0.9961\\
        DeiT-S & 0.9944 & 0.9864 & 0.9936\\
        Swinv2-S & 0.9944 & 0.9887 & 0.9940 \\
    \hline
    \end{tabular}
    \label{tab:result3}
\end{table}

\begin{table}[b]\small
    \caption{The test accuracy on the test dataset after 50 epochs re-training from checkpoints pre-trained on ImageNet-1K for each model.}
    \centering
    \begin{tabular}{c c   c c }
    \hline
        Test Accuracy    & P2 (2-class)  & P2 (3-class) & ACI (2-class)  \\
        \hline
        ResNet18 & 0.8950 &  0.7875 & 0.9320\\
        ViT-base & 0.9596 &  0.8970 & 0.9896\\
        DeiT-S & 0.9522 & 0.8901 & 0.9832\\
        Swinv2-S & 0.9520 & 0.9002 & 0.9870 \\
    \hline
    \end{tabular}
    \label{tab:result4}
\end{table}

\begin{figure}[!htbp]
\centering
\includegraphics[width=1\linewidth]{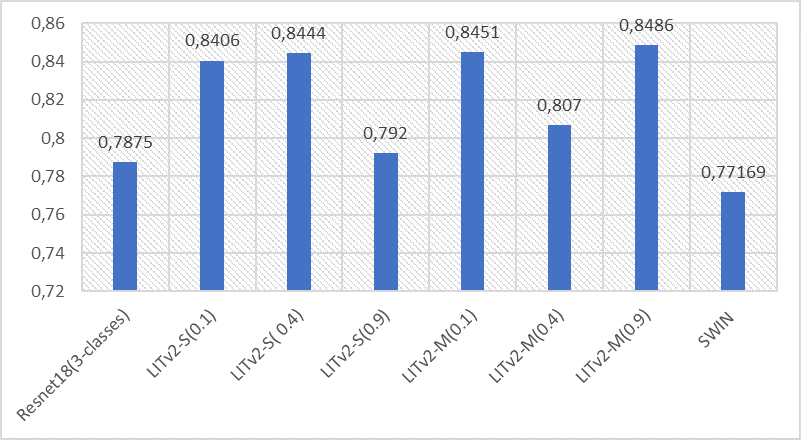}
\caption{Image classification accuracy with different models on the P2 3-class validation dataset with all models trained from scratch.}
\label{fig6}
\end{figure}

Additionally, we extended the comparison to evaluate the performance of ResNet18, Swin, LITv2-small, and LITv2-medium at different ratio values on the P2 3-class dataset, aiming to further analyze the impact of local and global attention on classification. The results are provided in Fig.~\ref{fig6}. 

Of all these models, the medium LITv2 model with an attention-head ratio rate of 0.9 achieved the highest overall accuracy, which is 84.86\%. In general, LITv2-based models trained directly on the new data tend to perform better than models, of which the overall accuracy is above 79.20\%. Additionally, LITv2 models, whether small or medium, achieved their best performance at a ratio of 0.4. This suggests that models incorporating both high-frequency and low-frequency attention tend to outperform those that overly emphasize local or global features. The LITv2 medium models have shown consistent and stable performance than the LITv2 small models at the cost of computation resources.

\begin{figure}[!htbp]
\centering
\includegraphics[width=1\linewidth]{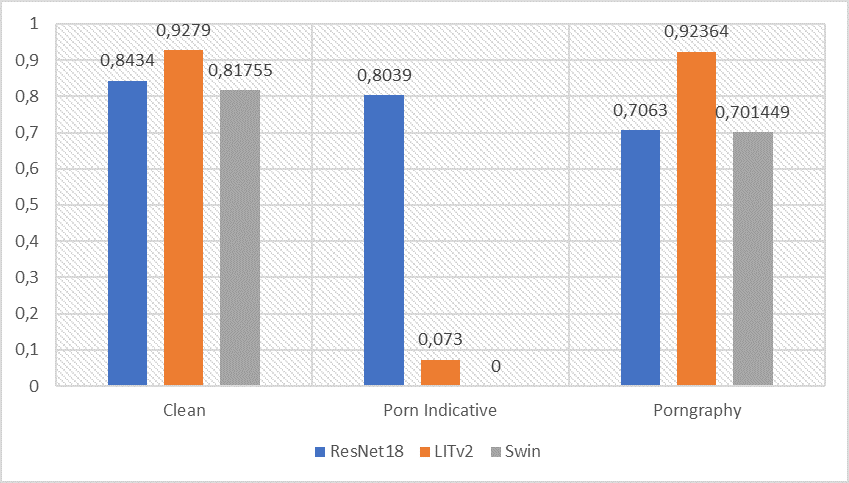}
\caption{Image classification per-class accuracy with different models on P2 3-class dataset}
\label{fig7}
\end{figure}

To gain further insights into the distinctions in category-wise classification results between the LITv2 model and other models, we compared the performance of the LITv2 medium model with a ratio value of 0.9 (as demonstrated in Fig. \ref{fig6}) with that of ResNet18 and Swin models. The results are illustrated in Fig. \ref{fig7}.

Compared to the other two ViT-based models, the ResNet pre-trained model achieved a more balanced performance over all three categories, especially in the indicative category. For the LITv2 model, which is trained on the new data set rather than Imagenet, the LITv2 medium model demonstrated higher recognition ability in clean and pornographic images than the other two models. Compared to the LITv2 medium model, the Swin model did not show an ideal classification result on the P2 3-class dataset which proves that the high-frequency and low-frequency structures can achieve a better balance of local features and global features.

\subsection{3-Class classification}
\label{sec:rationale}
\begin{figure}[!tbp]
\centering
\includegraphics[width=1\linewidth]{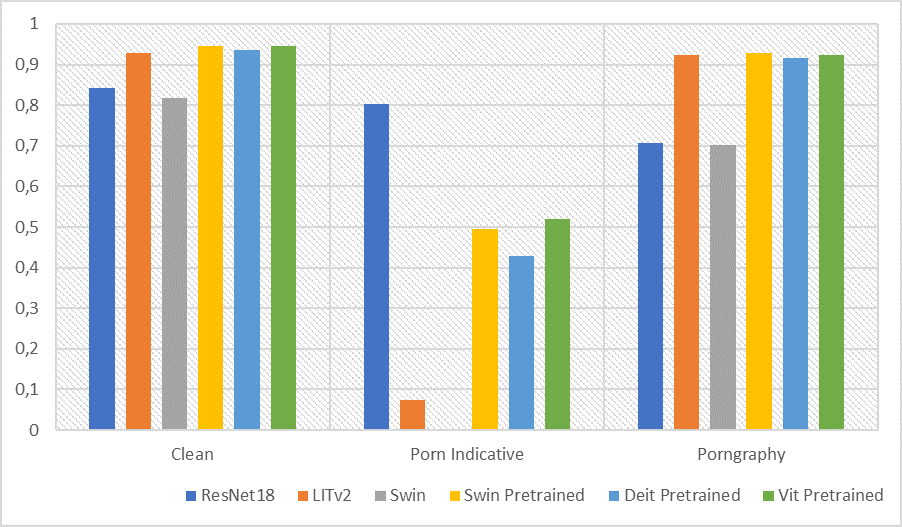}
\caption{Image classification accuracy with different models on the P2 3-class test dataset.}
\label{figs-1}
\end{figure}

To further compare the performance of different models in a three-class classification setting, we examined the outputs of various vision transformer models that have been fine-tuned on the ImageNet-1K dataset. We also compared these results with the classification results of other models completely trained from scratch in Fig. \ref{figs-1}. 

Through analysis, especially by comparing models trained from scratch and those fine-tuned models that were pre-trained on ImageNet-1K, it can be observed that there is a significant performance difference on the porn-indicative category classification. Based on pre-training, the Swin model outperforms ViT in the classification results for all three classes, while ViT surpasses DeiT. Additionally, even without pre-training, the LITv2 model demonstrates classification performance comparable to the three pre-trained models in the pornography and clean categories. Moreover, the introduction of the porn-indicative category does not significantly decrease the classification performance in the pornography and clean categories.

On the other hand, all models, except one fine-tuned ResNet model pre-trained on a large dataset that includes porn-indicative data, exhibit recognition accuracy in the porn-indicative category that are mostly below 50\%. This underscores the importance of adopting further penalty measures to ensure balanced classification results.

\section{CONCLUSION AND FUTURE WORK}
\label{sec:conclusion}
In this study, we conducted a comparison between the widely used ViT models, namely LITv2, ViT, DeiT, and Swin, and the pre-trained ResNet18 models for pornography image classification. We customized and fine-tuned these models with various configurations tailored to our specific datasets. We have compared models fine-tuned using checkpoints pre-trained on the ImageNet-1K data set with models trained from scratch across three datasets. We have also explored the performance differences among vision transformer models based on different attention principles. Additionally, we have examined variations in the performance of vision transformer models with different complexities and the impact of varying ratios of local attention to global attention. As a focal point of this paper, we delved into the variations in the classification performance of different models, particularly with the introduction of the ``porn-indicative" category in pornography images.

In general, models trained on more diverse training data (pre-trained models) tend to outperform models trained from scratch. The comparison results among vision transformers based on different attention principles indicate that the LITv2 model, compared to traditional ViT models, distillation-based DeiT models, and sliding window-based Swin models, demonstrates a better balance between high-frequency and low-frequency attention, leading to more effective image representation. Furthermore, in scenarios with sufficient computational resources, larger models with more parameters typically outperform smaller models. Compared to the classification of two-class pornography data, the fine-tuned models for the three-class pornography data showed minimal overfitting during training. Simultaneously, these models maintained high classification accuracy for both traditional pornography and clean categories.

A shortcoming of the paper is the limited number of images and unbalanced composition of datasets. To address this, we plan to incorporate transfer learning modules into our model to mitigate the dataset's limitations in the future. Another issue is the lack of ethically compliant sensitive image datasets. It is challenging to determine whether there is an overlap between datasets like ImageNet-1K and ACI, which could affect classification accuracy and results. We will create datasets that adhere to ethical standards for our research. To further analyze the impact of the vision transformer model on classification, we will incorporate more visualization methods to analyze classification results, addressing the challenge of presenting sensitive data more appropriately. Additionally, we plan to conduct a more extensive and fine-grained analysis of pornography images, exploring key details that determine explicit content.


\bibliographystyle{IEEEtran}
\balance
\bibliography{egbib}


\end{document}